%% file: main.tex
\documentclass[lettersize,journal]{IEEEtran}
\usepackage{amsmath,amsfonts}
\usepackage{algorithmic}
\usepackage{array}
\usepackage[caption=false,font=normalsize,labelfont=sf,textfont=sf]{subfig}
\usepackage{textcomp}
\usepackage{stfloats}
\usepackage{url}
\usepackage{verbatim}
\usepackage{graphicx}
\usepackage{cite}
\usepackage{algorithm}
\usepackage{multirow}
\usepackage{xcolor}
\usepackage{booktabs} 
\def\BibTeX{{\rm B\kern-.05em{\sc i\kern-.025em b}\kern-.08em
    T\kern-.1667em\lower.7ex\hbox{E}\kern-.125emX}}
\usepackage{balance}

\begin{document}
\title{Bitformer: An efficient Transformer with bitwise operation-based attention for Big Data Analytics at low-cost low-precision devices}
\author{\IEEEauthorblockN{Gaoxiang Duan\IEEEauthorrefmark{1}\IEEEauthorrefmark{2},
Junkai Zhang\IEEEauthorrefmark{1}\IEEEauthorrefmark{2},
Xiaoying Zheng\IEEEauthorrefmark{1}\IEEEauthorrefmark{2}\IEEEauthorrefmark{4}\thanks{\IEEEauthorrefmark{4} Corresponding authors: Xiaoying Zheng, Yongxin Zhu, Victor Chang},
Yongxin Zhu\IEEEauthorrefmark{1}\IEEEauthorrefmark{2}\IEEEauthorrefmark{4},
Victor Chang\IEEEauthorrefmark{3}\IEEEauthorrefmark{4}
}\\
\IEEEauthorblockA{\IEEEauthorrefmark{1}University of Chinese Academy of Sciences}\\
\IEEEauthorblockA{\IEEEauthorrefmark{2}Shanghai Advanced Research Institute, Chinese Academy of Sciences, Shanghai, China}\\
\IEEEauthorblockA{\IEEEauthorrefmark{3}Operations and Information Management, Aston Business School, Aston University, UK}

}

\maketitle

\begin{abstract}
In the current landscape of large models, the Transformer stands as a cornerstone, playing a pivotal role in shaping the trajectory of modern models. However, its application encounters challenges attributed to the substantial computational intricacies intrinsic to its attention mechanism. Moreover, its reliance on high-precision floating-point operations presents specific hurdles, particularly evident in computation-intensive scenarios such as edge computing environments. These environments, characterized by resource-constrained devices and a preference for lower precision, necessitate innovative solutions.

To tackle the exacting data processing demands posed by edge devices, we introduce the Bitformer model, an inventive extension of the Transformer paradigm. Central to this innovation is a novel attention mechanism that adeptly replaces conventional floating-point matrix multiplication with bitwise operations. This strategic substitution yields dual advantages. Not only does it maintain the attention mechanism's prowess in capturing intricate long-range information dependencies, but it also orchestrates a profound reduction in the computational complexity inherent in the attention operation. The transition from an $O(n^2d)$ complexity, typical of floating-point operations, to an $O(n^2T)$ complexity characterizing bitwise operations, substantiates this advantage. Notably, in this context, the parameter $T$ remains markedly smaller than the conventional dimensionality parameter $d$.

The Bitformer model in essence endeavors to reconcile the indomitable requirements of modern computing landscapes with the constraints posed by edge computing scenarios. By forging this innovative path, we bridge the gap between high-performing models and resource-scarce environments, thus unveiling a promising trajectory for further advancements in the field.
\end{abstract}

\input{1.Intro}

\input{2.Prior_knowledge_Related_work}

\input{3.Method}

\input{4.Experiments}

\input{5.Conclusion}

\bibliographystyle{IEEEtran}
\bibliography{ref}

\input{Appendix}

\end{document}

%% file: 1.Intro.tex
\section{Introduction}
\IEEEPARstart{I}n recent times, the significant advancements in the field of deep learning particularly with Large Models (LMs) have brought about a huge revolution. LMs such as GPT\cite{GPT3} and SAM\cite{kirillov2023segment} exhibit exceptional capabilities in handling various tasks\cite{vu2016geosocialbound, khan2018real,9034081}which can comprehend and generate language that closely resembles human expression, as well as in interpreting diverse styles of images. The fundamental model supporting these LMs is the Transformer\cite{Transformer}.
The attention mechanism, as the most crucial feature extraction mechanism in the Transformer, performs information extraction by comparing tokens in the input sequence pairwise. Its advantages lie in its ability to generalize and capture long-range dependencies.

However, along with the increase in computational scale, LMs come with not only remarkable performance improvements but also significant challenges related to increased computational requirements and energy consumption. According to statistics, the training process of models like GPT-3\cite{GPT3} consumes electricity equivalent to the usage of 5000 typical households in a year\cite{desislavov2021compute}. In particular, in Transformer-based LMs, the attention mechanism stands out as the most computationally intensive component, necessitating $O(n^2d)$ complexity of floating-point matrix multiplication. This accounts for approximately $50\%$ of the entire inference process in a standard Transformer\cite{Transformer}.The increasing utilization of LMs necessitates a heightened focus on the energy and computing power consumed during both model training and inference. 
Furthermore, there is a strong need to place the LMs on edge devices. In the context of edge devices with limited power and computational capabilities, the task of processing and analyzing big data presents both a promising frontier and a formidable challenge. To leverage LMs for data analytics, the conventional approach often involves uploading data to central servers. However, if we can migrate the inference and even training processes to these edge devices, 
we can ensure user privacy by not uploading raw data while still enjoying excellent experiences.
Moreover, it would alleviate the demand for network bandwidth, reduce computational pressure on servers during peak times, and enhance real-time data analytics. One crucial initial phase in the implementation of LMs on edge devices involves the optimization of power and energy consumption, particularly with respect to the attention mechanism employed by Transformer.


To address the problem of high computational cost caused by float multiplication in the Transformer and the limited computation ability and power in edge devices, we propose Bitformer which uses a novel attention with bitwise operations to highly simplify the computation need of the Transformer while containing the ability of global range feature extraction. The core idea of our work is to replace the float matrix multiplication in attention with XOR operations on matrix of binaries. Using a brand new Time-Integrate-and-Fire(TIF) operation inspired by Spike Neural Network(SNN)\cite{roy2019towards} Integrate-and-Fire operation, the input float data is converted into $T$ time steps of binary data. Then in the field of binary format data, the attention score is obtained through XOR operation instead of the inner product operation. The final attention matrix that also contains similarity information using a more computationally efficient mechanism. Our contributions can be summarized as follow:
\begin{itemize}
    \item First, we devise a novel attention mechanism based on XOR operations for binary data, which is the kernel operation of Transformer models. This innovative approach retains the capacity for global feature extraction while not only reducing the theoretical computational complexity from $O(n^2d)$ to $O(n^2T)$, where $T$ is significantly smaller than $d$, but also capitalizing on bitwise XOR operations rather than involving intricate floating-point multiplications and additions which yields heightened computational efficiency and diminished energy consumption.
    These modifications are crucial for edge devices that are limited in their energy resources and computational power.
    \item Second, we develop a novel approach for converting the attention mechanism's data format to enable our attention mechanism based on XOR operations. This conversion process entails representing input floating-point numbers as a composite of T binary fractions, all within a quantifiable margin of error. When executing attention operations on low-precision devices, this method strategically trades a marginal yet quantifiable level of precision for notable benefits. These advantages include a reduction in power consumption demands during calculations and a simultaneous enhancement in computational efficiency.
    \item Last, we implement and validate our proposed methods via multiple types of tasks with significant performance improvement. The experimental results demonstrate that Bitformer achieves comparable or higher performance in various tasks. Our approach achieves $90.2\%$ accuracy in social media text classification, which is higher than Transformer and $80.2\%$ in ImageNet classification. Remarkably, Bitformer achieves this performance while significantly reducing computational complexity compared to the standard Transformer. 
\end{itemize}

The remainder of the paper is organized as the following.
Section 2 delves into related work in the field, exploring notable contributions such as Efficient Transformer and attempts at edge computing for big data analytics.
Section 3 lays the groundwork by introducing essential background knowledge, including attention mechanism, Hamming distance, and SNN.
Section 4 presents our proposed method, highlighting data format conversion and the innovative attention mechanism we have developed.
Section 5 showcases the experimental evaluation of our approach across a range of language and image tasks.
Finally, we conclude by summarizing our findings.

%% file: 2.Prior_knowledge_Related_work.tex
\begin{figure*}[h]
    \centering
    \includegraphics[width = 0.9\linewidth]{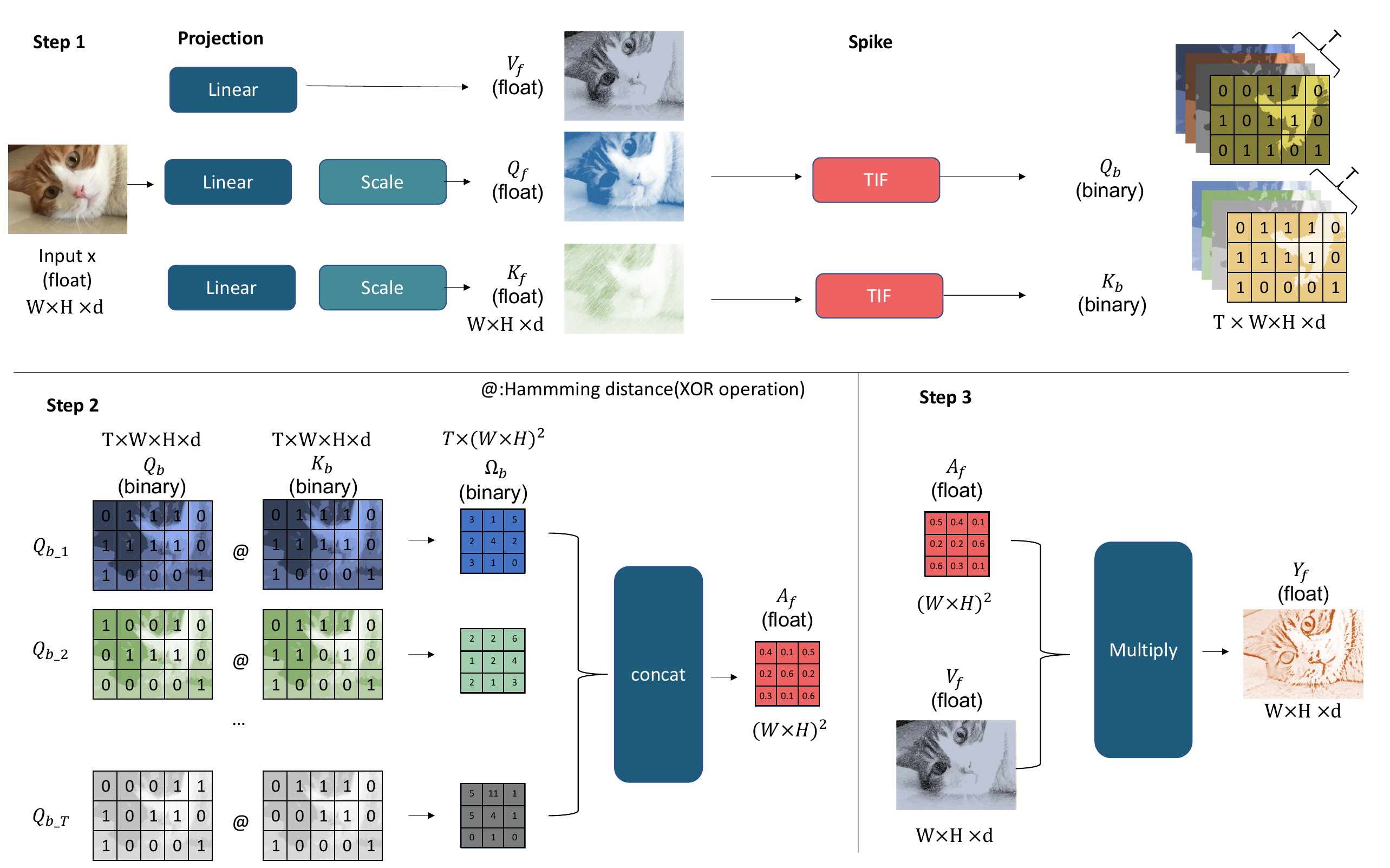}
    \caption{The Bitformer attention mechanism, which we illustrate using an image input, consists of three steps, aligning with the method description. Our bitwise attention mechanism incorporates two key ideas. In the first step (Step 1), we employ Time Integrate-and-Fire (TIF) to convert the float data $Q_f$ and $K_f$ into binary data $Q_b$ and $K_b$. This enables us to transform the attention operation into a binary operation. In the second step (Step 2), instead of utilizing dot-product, we utilize the Hamming distance to assess the similarity between each token, leveraging the XOR operation. By concatenating the distance scores for all time steps, we complete the attention operation and transition back to the real number field. Our approach can be applied to various domains, and here we exemplify its application using an image of size $W\times H\times d$ to illustrate its functionality.}
    \label{fig:bitformer}
\end{figure*}

\section{Related works}
\subsection{Transformer}
The Transformer model, introduced by Vaswani et al.\cite{Transformer}, has become a cornerstone in many fields, including the analytics of big  data\cite{9732646,9794444,9435103}. Its attention mechanism, which allows the model to weigh and prioritize different parts of the input sequence, has proven particularly effective in handling a variety of tasks. Several notable models, such as BERT\cite{bert}, GPT\cite{GPT3}, and others\cite{swin,biformer,reformer}, have been built upon the Transformer architecture, demonstrating remarkable success in data analytic tasks like text classification\cite{zhang2021fast,li2021act,wang2019learning}, translation\cite{raganato2018analysis,di2019adapting}, and image recognition\cite{vit,meng2022adavit}. These models have also been instrumental in analyzing big data, providing insights into data trends, user behavior, and other phenomena\cite{han2021transformer}. However, despite their effectiveness, these models are often computationally intensive, which makes them challenging to deploy on devices with limited computing power\cite{lin2022survey}. This is particularly relevant in the context of big data, where the volume, velocity, and variety of data can be overwhelming for traditional models.
\subsection{Efficient Attention Mechanism for Edge Devices}
The computational complexity of the Transformer model, particularly its attention mechanism, poses a significant challenge for deployment on edge devices, which often have limited computational resources\cite{pintus2012paraimpu}. Several research efforts have been dedicated to simplifying and optimizing the attention mechanism to make it more suitable for edge devices.
Various works\cite{reformer,blockbert,longformer,star,biformer} have proposed methods to reduce the computational load of the attention mechanism without significantly sacrificing the model's performance, with a particular focus on making these models more efficient for edge devices. The Reformer\cite{reformer}, for instance, classifies key lists by correlation and reorders them accordingly with a hashing operation. Attention is focused only within highly correlated chunks, reducing the computational demands and making it more suitable for edge devices.
Similarly, BlockBert\cite{blockbert} uses a different substitution function in its pre-processing to reduce computational load, while Biformer\cite{biformer} offers a bi-level routing attention mechanism which first filters query-key pairs at a coarse region level attention to remove irrelevant pairs then do a fine-grained attention with relative tokens. These methods aim to maintain the effectiveness of the attention mechanism while reducing the computational demands, making them more suitable for deployment on edge devices.
However, a common limitation of these methods is their reliance on float multiplication, which can be computationally expensive and may not be suitable for edge devices with limited computational power. Therefore, further research and innovation are needed to develop attention mechanisms that are both effective and efficient for deployment on edge devices.


\subsection{Spike Neuron Network}
The quick advancement of deep learning technology has created a greater need for artificial intelligence computing. As a result, individuals have started suggesting new devices, computer architectures, and deep learning frameworks.\cite{yu2021compute,dean20201,richards2019deep,rathi2020enabling} Among them, Spike Neural Network (SNN) as the third generation of neural networks has gained significant attention for their low power consumption and biological plausibility\cite{roy2019towards}. SNN differs from traditional deep learning models by employing spike sequences rather than continuous decimal values for information computation and transmission. This characteristic sets them apart in the realm of neural network architectures. Thanks to the advancements in artificial neural network (ANN), SNNs have been able to improve their performance by adopting advanced architectures from ANN\cite{hu2021spiking,hunsberger2015spiking,zheng2021going}. However, despite SNN's advantage of low power consumption, their model performance is still significantly lower than that of ANN models. This phenomenon remains present in SNN-based Transformer models\cite{zhou2022spikformer}.

\section{Prior knowledge}

\subsection{Attention Mechanism}
Attention is a pivotal component within the Transformer model, facilitating enhanced information exchange, capturing long-range dependencies, and enabling precise focus on relevant parts of the input sequence. We will introduce the implementation of the attention operation in three steps.
\begin{itemize}
    \item Step 1: For input $X$, we first perform linear projection on $X$ using three learnable linear matrices $W_q$,$W_k$,$W_v$, and obtain three float objects $Q_f$, $K_f$ and $V_f$:
    \begin{equation}
        Q_f = XW_q,K_f = XW_k, V_f = XW_v
    \end{equation}
    where $f$ denotes the format of the data is the float type.
    \item Step 2: The attention matrix $A$ is a measure of the relevance or similarity between the $Q_f$ and $K_f$. It is computed by calculating the dot product between the query and key, scaling it by the square root of the key vector dimension, and applying a softmax operation to normalize the values:
    \begin{equation}
       A = \text{attention}(Q_f,K_f)=\text{softmax}(\frac{Q_f K_f^T}{\sqrt{d} })
    \end{equation}
    where $d$ denotes the key vector dimension. Softmax operation ensures every attention value is positive.
    \item Step 3: The final step of the attention mechanism involves obtaining the output by linearly combining the values $V_f$ using the attention matrix obtained in the previous step:
    \begin{equation}
       Y_f= AV_f
    \end{equation}
\end{itemize}

Attention operation achieves feature extraction by pairwise comparisons, enabling the re-representation of all tokens in the input, thereby facilitating effective information aggregation and feature learning.
\subsection{Spike Neurons}
Spike neurons are the basic units in a SNN. They receive continuous input signals and integrate them over time. When the integrated signal reaches a certain threshold, they output discrete spikes, transitioning from continuous float values to binaries 0 and 1, enabling the representation and processing of temporal information in the network.

In the SNN, the Integrate-and-Fire (IF) node is one of the basic neuron node, simulating the basic dynamic behavior of neurons. In this node, the neuron receives input signals, integrates them, and fires a spike when the accumulated signal strength exceeds a preset threshold, after which the state is reset. The output $V$ after $\text{IF}$ can be expressed as a function of input $S$ and time step $T$:
\begin{equation}
    S = \text{IF}(X, T)
\end{equation}
In the IF spike neuron node, the membrane potential $H[t]$ at each time step $t$ is updated using the following equation:
\begin{equation}
    H[t] = V[t - 1] + X[t]
\end{equation}
where $X[t]$ represents the input current at time step $t$. When the membrane potential $H[t]$ exceeds the firing threshold $V_{th}$, the spike neuron generates a spike $S[t]$:
\begin{equation}
    S[t] = \Theta (H[t] - V_{th})
\end{equation}
where $\Theta(x)$ represents the Heaviside function:
\begin{equation}
\Theta(x)=
\begin{matrix} 
  1, x\ge 0 \\ 
  0, x <0
\end{matrix}.
\end{equation}
The membrane potential after the spike is triggered, denoted as $V[t]$, depends on whether a spike was generated or not. If no spike is generated, $V[t]$ remains equal to $H[t]$. However, if a spike is generated, then $V[t]$ is reset as:
\begin{equation}
    V[t] = H[t](1 - S[t]) + (H[t] - V_{th}) S[t]
\end{equation}

\subsection{Hamming Distance}

Hamming distance is a measure used to determine the degree of difference between two sequences of equal length. It does this by counting the number of different characters in the same position. In other words, the Hamming distance is the minimum number of substitutions needed to transform one sequence into another. It is a simple and intuitive calculation that finds wide use in the fields of information theory, computer science, and bioinformatics.

Calculating the Hamming distance is a straightforward process. You compare two sequences $a$, $b$ bit by bit and count the number of different characters they have in the same position:
\begin{equation}
      \phi(a,b)  = \sum_{i=1}^{d} \text{XOR}(a_i,b_i)
\end{equation}
XOR represents the exclusive or operation. In particular, for two matrices A and B, where each matrix consists of n d-dimensional vectors, we can obtain a matrix of Hamming distance that describes the distance relationship between the two pairs:
\begin{equation}
     D = \Phi(A, B)
\end{equation}
\begin{equation}
     D_{i,j} = \phi (A_i,B_j)
\end{equation}

Where $D_{i,j}$ represents the row $i$ and column $j$ of D.

\begin{figure*}
    \centering
    \includegraphics[width = \linewidth]{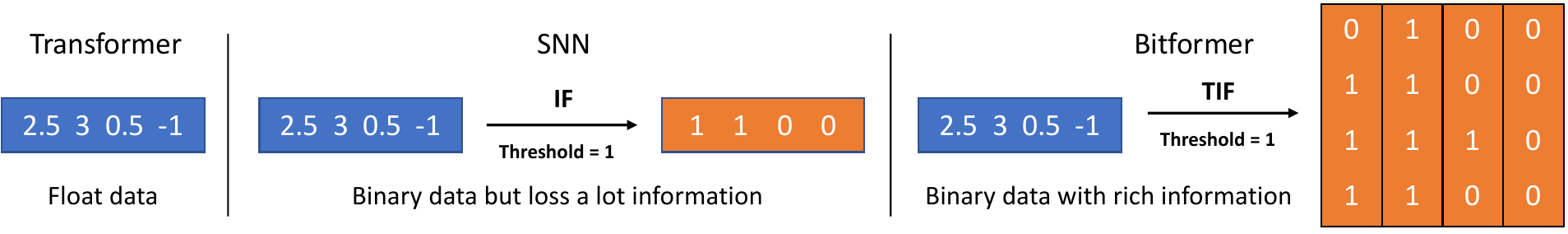}
    \caption{Comparing Hamming distance to Dot product for binary data. The standard float operations provide the richest information, but also require the most expensive compute consumption. On the other hand, SNN-based methods use spike operations like IF to convert data into binary space, which is power-friendly but lacks a lot of information. Our method involves converting a single float data into a combination of a series of binaries data, which reduces compute consumption while minimizing information loss.}
    \label{fig:format-compare}
\end{figure*}

\begin{figure*}
    \centering
    \includegraphics[width = \linewidth]{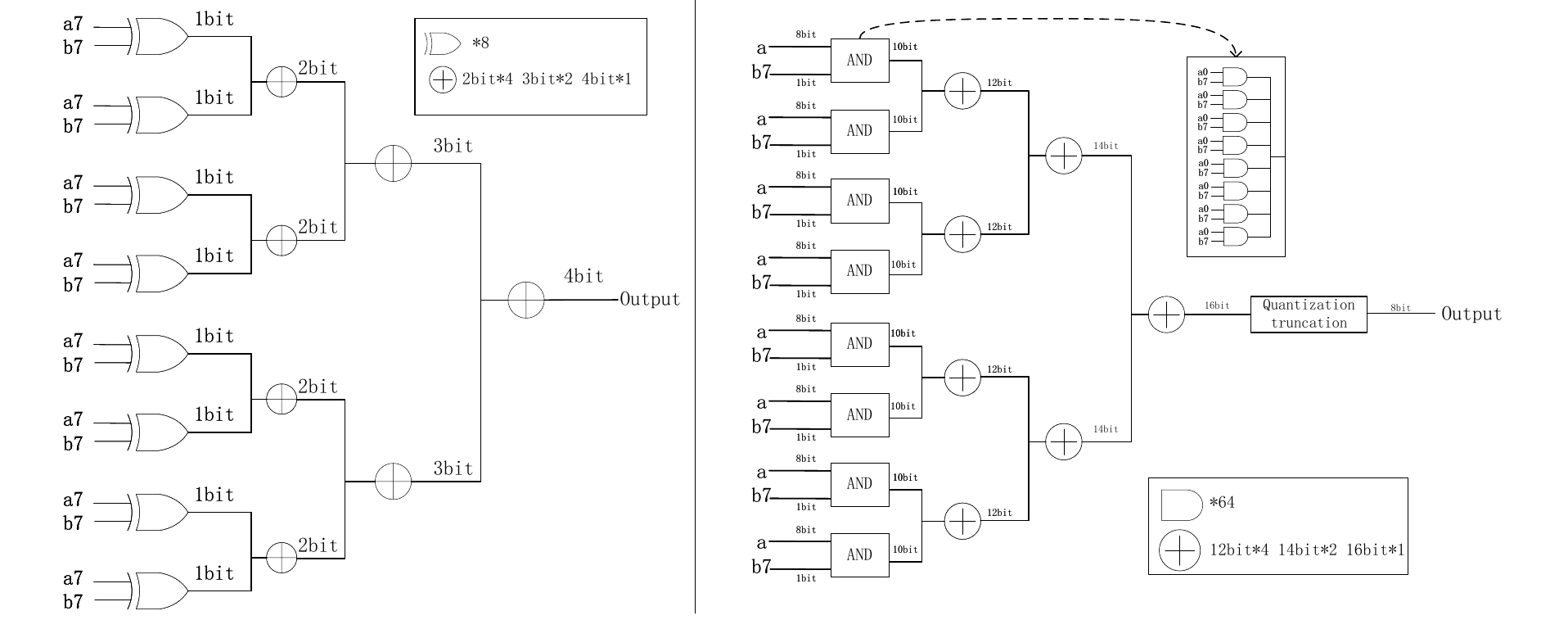}
    \caption{Comparing 8bit Hamming distance with 8bit Dot product in circuit level. }
    \label{fig:circuit}
\end{figure*}

%% file: 3.Method.tex
\section{Method}

We propose an edge device-friendly model called Bitformer, which consists of a novel data format conversion and a brand new attention mechanism based on bitwise operation. Our approach significantly reduces computational complexity while preserving performance. In Section 3.1, we present our Bitformer and provide a thorough overview of its implementation. In Section 3.2, we analyze the influence of data format conversion on data analytics. In Section 3.3, we analyze the brand new attention mechanism for binary. In Section 3.4, we analyze the complexity of the new attention mechanism.
\subsection{Bitformer Implementation}
An overview of Bitformer is depicted in Fig.\ref{fig:bitformer}. To facilitate comparison with the standard attention implementation, we will introduce our bitwise attention operation in three steps, as previously done.
\begin{itemize}
    \item Step1: Projection and Spike.\\
    Similar to the standard attention mechanism, we begin by performing a linear projection to obtain float values for $Q_f$, $K_f$, and $V_f$:
    \begin{equation}
        Q_f = XW_q,K_f = XW_k, V_f = XW_v
    \end{equation}
    Prior to performing spike operations on $Q_f$ and $K_f$ we apply preprocessing steps to these matrices.\\
    We modify the $Q_f$, $K_f$ using a linear transformation to ensure all values are positive and enable better information retention:
    \begin{equation}
        Q_f = \frac{Q_f - \min(Q_f)}{\max(Q_f)}
    \end{equation}
    \begin{equation}
       K_f = \frac{K_f - \min(K_f)}{\max(K_f)}
    \end{equation}

   We'll use a TIF (T time step IF) operation to complete the transition from $Q_f$ to $Q_b$, based on the IF node:
    \begin{equation}
       Q_b= \mathit{\text{TIF} } (Q_f, T) = \text{IF}(\left [X^1,X^2,..,X^T \right ], T)
    \end{equation}
   which $X^i = X$, $b$ represents the $Q_b$ is a binary type data.
    
    \item Step2: XOR Attention.\\
    We now have binaries matrix $Q_b$ and $K_b$. We use Hamming distance instead of performing a dot product operation to measure the similarity between the two on each time $t$. This will produce a sequence of attention matrices $A_t$: 
    \begin{equation}
        A_t = \Phi(Q_b^t, K_b^t) 
    \end{equation}
    Here, $Q_b^t$, $K_b^t$ represents the element in the time step $t$ of matrix $Q_b$ and $K_b$.
   
    Finally, we combine the Hamming distances of all T dimensions. The farther the distance, the lower the similarity. This allows us to obtain the final attention score:
    \begin{equation}
         A = \frac{1}{\sum_{t = 1}^{T}A_t  + 1 } = \frac{1}{\sum_{t = 1}^{T}\Phi(Q_b^t, K_b^t)  + 1}
    \end{equation}
    We add $1$ to the denominator to prevent extreme cases where the distance is calculated as $0$. The pseudo-code of bitwise attention is shown in Algorithm 1.
   
    \item Step3: Our third step is identical to the standard attention operation:
    \begin{equation}
       Y_f= AV_f
    \end{equation}
Once we obtain the attention matrix, we multiply it with V to obtain our final output. The final output is also a float-type matrix.
\end{itemize}

\begin{algorithm}
    \caption{TIF Conversion Algorithm}
    \label{alg:Bitformer}
    \begin{algorithmic}[1]
    \STATE Input $Q_f, K_f, V_f$ with size [n, d]
    \STATE \textbf{Pre-Transform}
    \STATE Transformation $Q_f$ and $K_f$:\\
    $Q_f = (Q_f - \min(Q_f))/\max(Q_f)$ \\
    $K_f = (K_f - \min(K_f))/\max(K_f)$ 
    \STATE \textbf{Time-Integrate-and-Fire}
    \STATE Initialize $Q_b$, $K_b$, $H$, $V$ with zeros with size [T, n, d]:\\
    $Q_b = [Q_b^1, Q_b^2,..., Q_b^T]$\\
   
    \STATE Convert $Q_f$ and $K_f$ to binary data: \\$Q_b = \text{TIF}(Q_f)$, $K_b = \text{TIF}(K_f)$\\
    \FOR{$t$ in Time step T}
        \STATE$H_q^t = V_q^{t-1} + Q_f$ 
        \STATE $Q_b^t = \Theta (H[t] - V_{th})$ 
        \STATE$V[t] = H[t](1 - Q_b^t) + (H[t] - V_{th}) Q_b^t$
    \ENDFOR\\
    \STATE\COMMENT{the calculation of $K_b$ is the same}\\
    \STATE\textbf{Output}: {$Q_b$, $K_b$ (binary matrices with size [T, n, d])}
    \end{algorithmic}
 \end{algorithm}

\begin{algorithm}
    \caption{Bitwise Attention Algorithm}
    \label{alg:Bitformer2}
    \begin{algorithmic}[1]
    \STATE Input $Q_b,K_b,V_b $  with size [T, n, d]
    \STATE Initialize $A_t$ with zeros with size [n,n]
    \STATE Compute Hamming distance matrix: 
    \FOR{$t$ in Time step T}
        \FOR{$i$ in column n}
            \FOR{$j$ in row n}
                 \STATE$A_{i,j}^t = \sum_{i=1}^{d} \text{XOR}(a_i,b_i)$
            \ENDFOR
        \ENDFOR
    \ENDFOR
    \STATE Compute Attention matrix A with size [n,n]: 
    \STATE$A = 1/(\sum(A_i) + 1)$\\
    \STATE \textbf{output}:{$A$ (float point matrix)}
    \end{algorithmic}
\end{algorithm}

\subsection{Bitformer V.S. Quntification}
A common method to reduce the computational resource consumption of Transformer is to quantize data and weights. Quantizing 32-bit float point numbers to 8 bits makes them comparable in size to the data with a time step of 8 in Bitformer. However, there is still a significant difference in computational resource consumption between two 8-bit fixed-point multiplications and XOR operations with 2-bit binary numbers. Fig.\ref{fig:circuit} illustrates the circuit implementations of classic tree-structured 8-bit Hamming distance and multiplication operations. It can be observed that the number of computing units and adder scales used in the two operations differ significantly. Considering that basic operations like attention mechanisms are computed $n^2d$ times, this performance gap becomes even more significant.

Despite numerous optimizations in existing hardware implementations for fundamental operations like multiplication, there is still a performance advantage in computing Hamming distance. We used High-Level Synthesis (HLS) on Xilinx xxx to compare the performance and resource consumption of Hamming distance and multiplication, as shown in Figure 11. The results indicate xxx.

\subsection{Data Format Conversion}
For edge devices constrained by computational power, a pivotal strategy for simplification involves the manipulation of data types. Altering from high-precision, storage-intensive floating-point types to fundamental binary data types that align with the capabilities of the device substantially alleviates both computational and storage burdens.

However, it is imperative to acknowledge that a decrease in data precision can potentially result in the loss of crucial information. Our subsequent experiments underscore this, revealing that information loss can yield pronounced performance degradation in data analytics tasks, especially evident in SNN-based models\cite{zhou2022spikformer} when confronted with intricate challenges. Such an outcome is incongruent with the requisites of effective big data analytics.

In scenarios where the simplification of operations engenders a notable impact on data analytics performance and fails to faithfully underpin subsequent recommendations and insights, the pursuit of data analytics on edge devices becomes futile. To circumvent this predicament, we have implemented two measures aimed at ensuring that the conversion of data into binary form can streamline computations without compromising performance integrity.

Firstly, the binary data format is exclusively maintained during the attention operation, with the input and output retaining their floating-point nature. Secondly, diverging from the direct conversion methodology employed by SNN-based models, we opt to transform a floating-point datum into a composite of T binary data units. The cumulative sum of these T binary data units approximates the quantifiable error inherent in the original floating-point datum (as demonstrated in the appendix A). Given that the final attention calculation also entails summing the outcomes of these T data units, this approach serves to curtail data information loss to a minimum.

Additionally, owing to the intrinsic attributes of the data type, the memory usage of T binary data will not be T times larger than that of a single floating-point data.

\subsection{Bitwise Attention}
When implementing attention operations in the context of edge computation, a intuitive approach might be to directly adapt the conventional attention mechanisms. However, such an approach proves neither resource-efficient nor computationally effective. Firstly, it would still necessitate binary matrix multiplications of the same dimensions as before, along with subsequent softmax operations and both of which impose substantial computational demands. Secondly, while the primary objective of attention remains consistent in capturing pairwise similarities for enhanced feature extraction, the conventional inner product measure effective for floating-point data does not seamlessly extend to the binary domain. Specifically, zero bits in binary data fail to discern differences in other vectors at the corresponding bit positions.

To counter these challenges, we propose a bitwise based attention operation for binary data within the context of big data analytics. We employ the Hamming distance as a measure of similarity between two binary vectors. This approach not only effectively captures similarity but also leverages XOR operations, which are more computationally friendly than multiplication. Furthermore, since all computed distances are non-negative, we can eliminate the need for softmax operations, thereby further reducing computational costs.

In summary, our bitwise attention mechanism leverages the strengths of Hamming distance and XOR operations to streamline attention computations on binary data, thereby reconciling the intricacies of binary representations with the demands of efficient edge computation in big data analytics.
\subsection{Complexity}
The traditional attention mechanism computes a dot product for each pair of $Q_f$ and $K_f$, which requires $O(n^2d)$ operations, where $n$ is the number of the token and $d$ is the dimension of the token. In contrast, the proposed method computes the Hamming distance for each pair of $Q_b$ and $K_b$, which can be done in $O(n^2)$ time. However, this needs to be done $T$ times, so the overall computational complexity is $O(n^2T)$. In Tab.\ref{tab:compare_complexity}, we assessed the complexity of the Transformer model, the latest efficient Transformer methods, and our own method. 

\begin{table*}[]
\centering
\caption{Comparation of complexity}
\label{tab:compare_complexity}
\resizebox{0.9\linewidth}{!}{
\begin{tabular}{lllll}
\hline
\textbf{Model}                                 & \textbf{Time Complexity}                           & \textbf{Data Type} & \textbf{Campare Similiarity} & \textbf{Compute Range} \\ \hline
Transformer\cite{Transformer} & $O(n^2d)$                           & Float              & Dot production               & All n tokens              \\
Reformer\cite{reformer}       & $O(2nd)+O(n^2d/4)$               & Float              & Dot production + Hashing     & n/4(in the same hash bucket)                     \\
BlockBert\cite{blockbert}     & $O(n^2d)+(n^2d/4)$ & Float              & Dot production               & n/4(in the same block)                       \\
Star Transformer\cite{star}   & $O(6nd)$                                             & Float              & Dot production + Location    & 5 ajacent + 1 global      \\
Bitformer                                      & $O(Tn^2)$                           & Binaries           & XOR                          & All n tokens              \\ \hline
\end{tabular}}
\end{table*}

\begin{table}[]

\centering
 \caption{Accuracy on CIFAR}
 \label{tab:Acc-CIFAR}
\begin{tabular}{lcccl}
\hline
\textbf{Model}     & \textbf{Type}        & \textbf{Time Step} & \textbf{\begin{tabular}[c]{@{}c@{}}CIFA10 \\ acc(\%)\end{tabular}} & \textbf{\begin{tabular}[c]{@{}l@{}}CIFAR100 \\ acc(\%)\end{tabular}} \\ \hline
VIT\cite{vit}        & ANN                  & -                  & 96.70                                                              & 81.02                                                                \\ \hline
Hybrid Training\cite{rathi2020enabling}    & \multirow{4}{*}{SNN} & 125                & 92.22                                                              & 67.87                                                                \\ \cline{1-1} \cline{3-5} 
STBP-tdBN\cite{zheng2021going}          &                      & 6                  & 92.92                                                              & 70.86                                                                \\ \cline{1-1} \cline{3-5} 
Spikeformer\cite{zhou2022spikformer}        &                      & 4                  & 93.94                                                              & 75.96                                                                \\ \cline{1-1} \cline{3-5} 
\textbf{Bitformer} &                      & 8                  & \textbf{95.88}                                                     & \textbf{80.13}                                                       \\ \hline
\end{tabular}

\end{table}

%% file: 4.Experiments.tex
\section{Experiments}
We conducted tests on common issues encountered in big data processing, which included tasks such as text classification and translation in NLP and image classification in CV.

We benchmarked our work against Transformer. Aside from using our original attention mechanism, we align our model with the target Transformer in terms of model layers, optimizer, hyperparameters, and other relevant aspects. All our experiments were conducted using 4 Tesla V100 GPUs.

To evaluate the performance and computation of our model, we chose some latest efficient Transformer works and SNN works for performance comparison. There will not show the performance of SNN works in NLP tasks since the compared SNN-based models did not provide implementations for NLP tasks.

\begin{table*}[]
\centering

\caption{Accuracy on Imagenet}
\label{tab:Acc-imagenet}
\resizebox{0.9\linewidth}{!}{
\begin{tabular}{lcccccc}
\hline
\textbf{Model}     & \textbf{Type}        & \multicolumn{1}{l}{\textbf{Architecture}} & \textbf{FLOPs/SOPs(G)} & \textbf{Params(M)} & \textbf{Time Step} & \textbf{Top1 acc $\%$} \\ \hline
Swin-T\cite{swin}             & \multirow{8}{*}{ANN} & \multirow{8}{*}{Transformer}              & 4.5            & 29              & -                  & 80.9              \\ \cline{1-1} \cline{4-7} 
CSWin-T\cite{dong2022cswin}            &                      &                                           & 4.5            & 23              & -                  & 81.8              \\ \cline{1-1} \cline{4-7} 
DAT-T\cite{xia2022vision}              &                      &                                           & 4.6            & 29              & -                  & 81.9              \\ \cline{1-1} \cline{4-7} 
CrossFormer-S\cite{wang2023crossformer}     &                      &                                           & 5.3            & 31              & -                  & 81.7              \\ \cline{1-1} \cline{4-7} 
RegionViT-S\cite{chen2021regionvit}        &                      &                                           & 5.3            & 31              & -                  & 81.6              \\ \cline{1-1} \cline{4-7} 
MaxViT-T\cite{tu2022maxvit}          &                      &                                           & 5.6            & 31              & -                  & 82.9              \\ \cline{1-1} \cline{4-7} 
ScalableViT-S\cite{yang2022scalablevit}      &                      &                                           & 4.2            & 32              & -                  & 82.8              \\ \hline
Spiking ResNet\cite{hu2021spiking}     & \multirow{4}{*}{SNN} & ResNet34                                  & 65.3           & 21              & 350                & 70.6              \\ \cline{1-1} \cline{3-7} 
STBP-tdBN\cite{zheng2021going}          &                      & ResNet34                                  & 6.5            & 22              & 6                  & 63.5             \\ \cline{1-1} \cline{3-7} 
Spikeformer\cite{zhou2022spikformer}        &                      & Transformer                               & 6.8            & 66.3            & 4                  & 70.1              \\ \cline{1-1} \cline{3-7} 
\textbf{Bitformer} &                      & Transformer                               & 3.1            & 22            & 8                  & \textbf{80.2}     \\ \hline
\end{tabular}
}
\end{table*}

\begin{table}[]
\centering
\caption{Text classification result on IMDB and THUCNews datasets}
\begin{tabular}{lccc}
\hline
\multirow{2}{*}{Dataset} & \multicolumn{3}{c}{Acc(\%)}                \\
                         & Transformer & BiLSTM & Bitformer \\ \hline
IMDB                     & 82.5        & 86.0             & 85.1      \\
Sports                   & 98.9        & 98.5             & 98.8      \\
Entertainment            & 95.4        & 95.2             & 94.5      \\
Home                     & 89.4        & 82.8             & 87.2      \\
Lottery                  & 80.2        & 81.9             & \textbf{87.1}      \\
Real Estate              & 96.8        & 97.2             & 93.1      \\
Education                & 90.6        & 89.5             & \textbf{95.5}      \\
Fashion                  & 88.9        & 80.5             & 85.1      \\
Politics                 & 82.0        & 81.6             & \textbf{88.5}      \\
Constellation            & 83.1        & 83.2             & \textbf{85.2}      \\
Game                     & 93.2        & 87.7             & 92.5      \\
Society                  & 85.4        & 82.8             & \textbf{85.7}      \\
Technology               & 93.7        & 88.4             & \textbf{96.6}      \\
Stock                    & 92.1        & 86.3             & \textbf{92.3}      \\
Finance                  & 83.2        & 78.6             & \textbf{85.1}      \\ \hline
Average                  & 89.0        & 86.8             & \textbf{90.2}
\\\hline
\end{tabular}
\label{tab:Text-classification}
\end{table}

 \begin{table}[]
    \setlength{\tabcolsep}{10mm}
    \centering
    \caption{Bleu scores of different models.}
    \begin{tabular}{ll}
    \toprule
    \hline
    Model             & Bleu Score     \\
    \hline
    Transformer\cite{Transformer}       & 25.8           \\
    BlockBert\cite{blockbert}     & 23.1             \\
    Reformer\cite{reformer}          & 22.3           \\
    Star-Transformer\cite{star}  & 20.1           \\
    \textbf{Bitformer} & \textbf{25.0}  \\
    \hline
    \bottomrule
    \label{tab:bleu-score}
    \end{tabular}
\end{table}

\subsection{Text Classification}
Text classification is a pivotal task in NLP and plays a crucial role in processing big data. To assess the efficacy of our model in handling big data, we selected two representative datasets.

The first dataset, THUCNews\cite{THUCTC}, is a compilation of approximately 740,000 news documents categorized into 14 groups. This dataset derived by filtering historical data from the Sina News RSS subscription channel is a rich source of user-focused content, mirroring the kind of data that circulates on social media platforms. 
Our second dataset is IMDB\cite{IMDB} which is a well-established benchmark in the NLP field. It comprises movie reviews with associated binary sentiment labels, indicating whether the review is positive or negative. 

As part of our training process, we established the number of training epochs to be 200 and set the learning rate to 0.0001. We used a batch size of 32 and opted to use Adam as the optimizer and utilized the glove.6B word embeddings. During training, we did not modify the embedding parameters.

We analyzed the performance of Bitformer in comparison with the original Transformer and BiLSTM\cite{Bilstm}. The outcomes on the THUCNews and IMDB datasets are displayed in Tab.\ref{tab:Text-classification}.
Based on our experimental results, we were able to achieve better performance than the two commonly used baseline models, Transformer and BiLSTM, in 8 out of 15 tasks. In fact, Bitformer was able to achieve a noteworthy 1.2-point improvement over the standard Transformer in the final average result.


\subsection{Translation}
The translation is a common NLP task and is widely used in big data processing to help with interaction across languages. Compared to classification, translation requires more accurate information extraction. We chose the WMT 2018 Chinese-English track focusing only on news articles as the translation dataset.

 Our training process consisted of 100 epochs, with the first epoch dedicated to warming up. To set the parameters, the Bitformer implementation is based on the model by Harvard\cite{opennmt}. We train the model with NoamOpt and without label smoothing.

Due to the lack of implementation of the SNN base model in text tasks, simply transplanting the model resulted in poor performance and lacked comparative value. Therefore, we selected other latest efficient Transformer implementations for comparison. The result shows in Tab.\ref{tab:bleu-score}.

The results showed that Bitformer's BLEU score reached 25, exceeding other efficient Transformer models and differing by only 0.8 from the standard Transformer. The results indicate that our updated attention implementation performs well, even when extracting more precise information, outperforming other efficient Transformers.

\subsection{Image Classification on CIFAR}
CIFAR10 is a dataset that comprises 60,000 32x32 color images, categorized into 10 classes with 6,000 images in each class. The dataset is split into 50,000 training images and 10,000 testing images. On the other hand, CIFAR100 presents a more complex classification problem with 100 classes, each containing 600 images, 500 training images and 100 testing images per class. When using Top1 accuracy as the evaluation metric, CIFAR100 is more challenging to classify than CIFAR10.

We have developed Bitformer based on the foundation of Vit Small\cite{vit} . The training process consists of 400 epochs, during which the images are divided into 16 patches of 4$\times$4.

Our model achieved an accuracy of 95.88\% in CIFAR10 TOP 1 accuracy, which is the highest among other SNN-based models. The difference is even more significant in CIFAR100, where our accuracy reaches 80.13\%. This is significantly higher than other models by at least 8 points and only 0.89\% lower than VIT, which shows the ability to replace the Transformer. The results of the experiment are shown in Tab.\ref{tab:Acc-CIFAR}.

\subsection{Image Classification on ImageNet}
The ImageNet dataset is highly regarded and influential in the field of computer vision. It contains approximately 1.3 million labeled images across 1,000 object categories. The input image size is set to 224$\times$224.

The training process consists of 300 epochs, with a warm-up period of 5 epochs. The batch size is 128, and the optimization algorithm employed is AdamW. A weight decay of 0.05 is applied to regularize the model. The learning rate schedule follows a cosine decay, and the initial learning rate is set to 0.001.

We categorized the results based on model type. The results are presented in Tab.\ref{tab:Acc-imagenet}. The results indicate that the performance of other SNN-based models is considerably lower than that of the general ANN model. Our Bitformer achieved a similar performance to that of mainstream Transformer models. In comparison to Swin-T\cite{swin}, Bitformer only differs by 0.7 percentage points, whereas the accuracy of other SNN-based models is at least 10 points lower.

In addition, it is worth noting that we generally use FLOPs to quantify the number of operations in a model, which refers to floating point operations. FLOPs represent the number of multiply-and-accumulate (MAC) operations. However, SNN-based models operate on binary data. Therefore, we use SOPs to measure the number of operations in the model, specifically representing binary accumulate (AC) operations. Different operations also have varying energy consumption. Based on the paper's assumption\cite{kundu2021hire} that MAC and AC operations are implemented on 45nm hardware, we have the following power consumption calculation for typical ANN models $a$:
\begin{equation}
    \mathcal{P}(a) = \mathit{E_{\text{MAC}}}\times \text{FLOPs}(a) = 4.6\mathit{pJ}\times  \text{FLOPs}(a)
\end{equation}

\begin{figure}
    \centering
    \includegraphics[width = 0.9\linewidth]{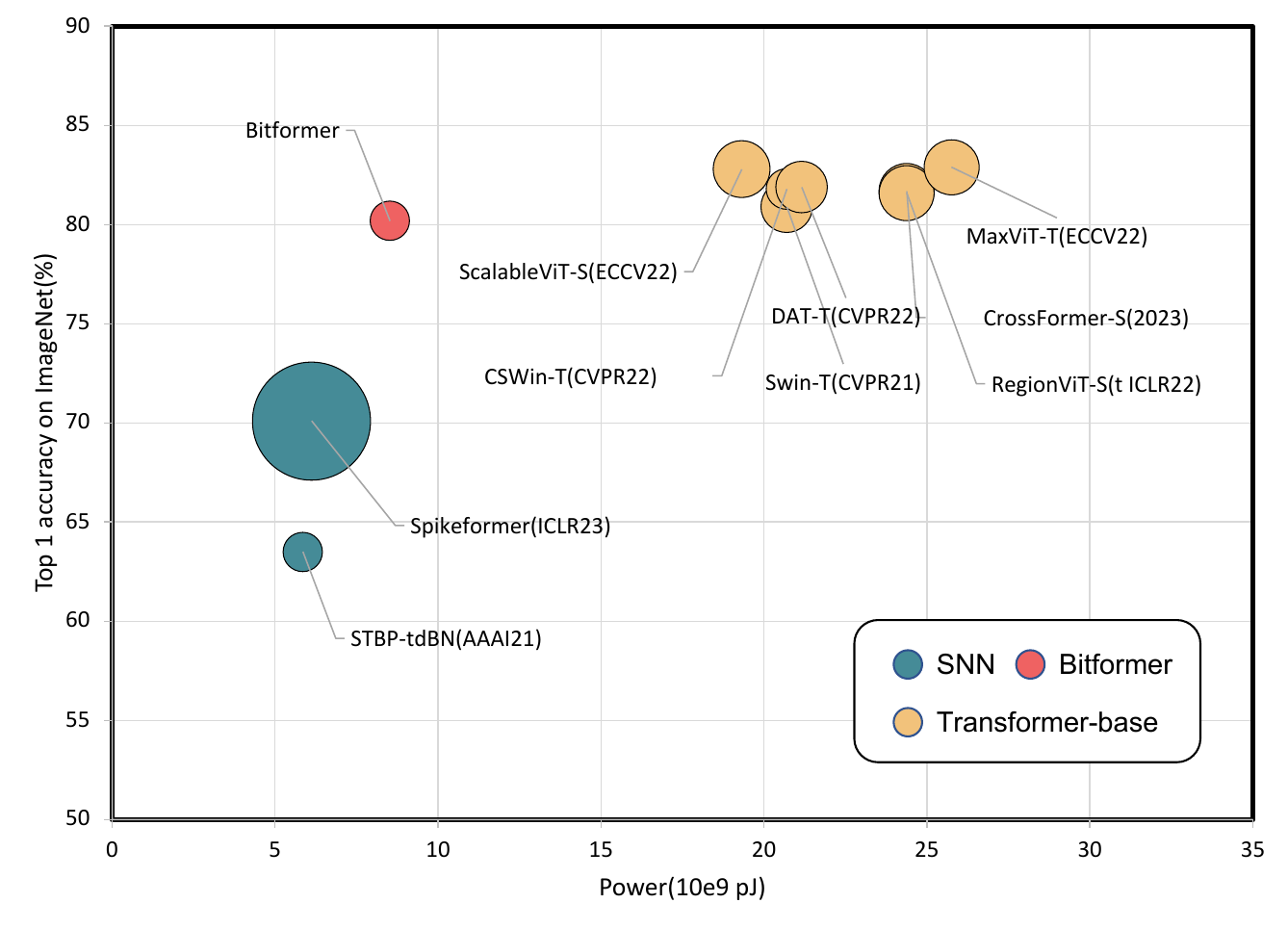}
    \caption{Compare the power and accuracy on ImageNet. Our method accomplished a well trade-off between power and accuracy.}
    \label{fig:scatter}
\end{figure}

For SNN-based models $b$, the power consumption can be calculated as:
\begin{equation}
    \mathcal{P}(b) = \mathit{E_{\text{AC}}}\times \text{SOPs}(a) = 0.9\mathit{pJ}\times \text{SOPs}(b)
\end{equation}

our approach can be viewed as an ANN-SNN mixed model. We use SOPs to calculate energy consumption for operations related to binary data in the attention layer, while using FLOPs for other parts. The final results are shown in Fig.\ref{fig:scatter}. 

In the scatter plot, each circle represents a model, and the area of the circle represents the model's parameter count. Detailed information regarding accuracy and parameters is provided in Tab.\ref{tab:Acc-imagenet}.

From the graph, we can clearly observe that SNN-based models generally have lower power consumption but struggle to achieve high accuracy. On the other hand, ANN-based models with higher accuracy tend to exhibit higher power consumption. Our model, however, demonstrates significantly lower power consumption compared to ANN-based models while achieving significantly higher accuracy than SNN-based models.

\subsection{Ablation Study on Time Step T}
We tested the impact of different time steps (T) on the performance of Bitformer on the CIFAR dataset, and the results are shown in Table \ref{tab:ablation}. From the results, it can be observed that, overall, a larger time step (T) leads to better performance.
\begin{table}[]
\caption{Ablation Study on Time Step T}
\begin{tabular}{lll}
\hline
Architecture design & CIFAR10 Top1 acc (\%) & CIFAR100 Top1 acc(\%) \\ \hline
ViT-S(32 float point)             & 96.7                  & 81.02                 \\
Bitformer T=2      & 69.32                 & 56.93                 \\
Bitformer T=4      & 82.24                 & 74.18                 \\
Biformer T=8      & 95.88                 & 80.13                 \\
Bitformer T=16     & 96.01                 & 80.55                 \\ \hline
\end{tabular}
\label{tab:ablation}
\end{table}

\subsection{Performance on FPGA}
The comparative analytics revealed that the Bitformer algorithm outperforms the Transformer algorithm on the Xilinx ZCU104 FPGA in several key aspects. This experimental setup aligns closely with the real-world application of our model in the edge device environment, where edge devices with limited computational resources are commonly employed. Fig.\ref{fig:FPGA} shows the results. Despite consuming 3.4\% more BRAM resources, Bitformer demonstrated a significant 14.5\% reduction in latency, indicating a faster processing time. Furthermore, it was substantially more efficient in terms of DSP, FF, and LUT usage, consuming 70.4\% less DSP and being 40.9\% and 36.4\% more efficient in FF and LUT usage, respectively. These metrics underline Bitformer's superior efficiency and speed, which are critical for processing large-scale data on edge devices. These results provide strong empirical evidence that Bitformer is more hardware-friendly, making it an ideal choice for deployment in edge devices with limited resources. These attributes significantly enhance the potential for efficient, real-time  data analytics in edge device environments, thereby bringing us one step closer to achieving the goal of efficient and effective big  data analytics at the edge.

%% file: 5.Conclusion.tex
\begin{figure}
    \centering
    \includegraphics[width = \linewidth]{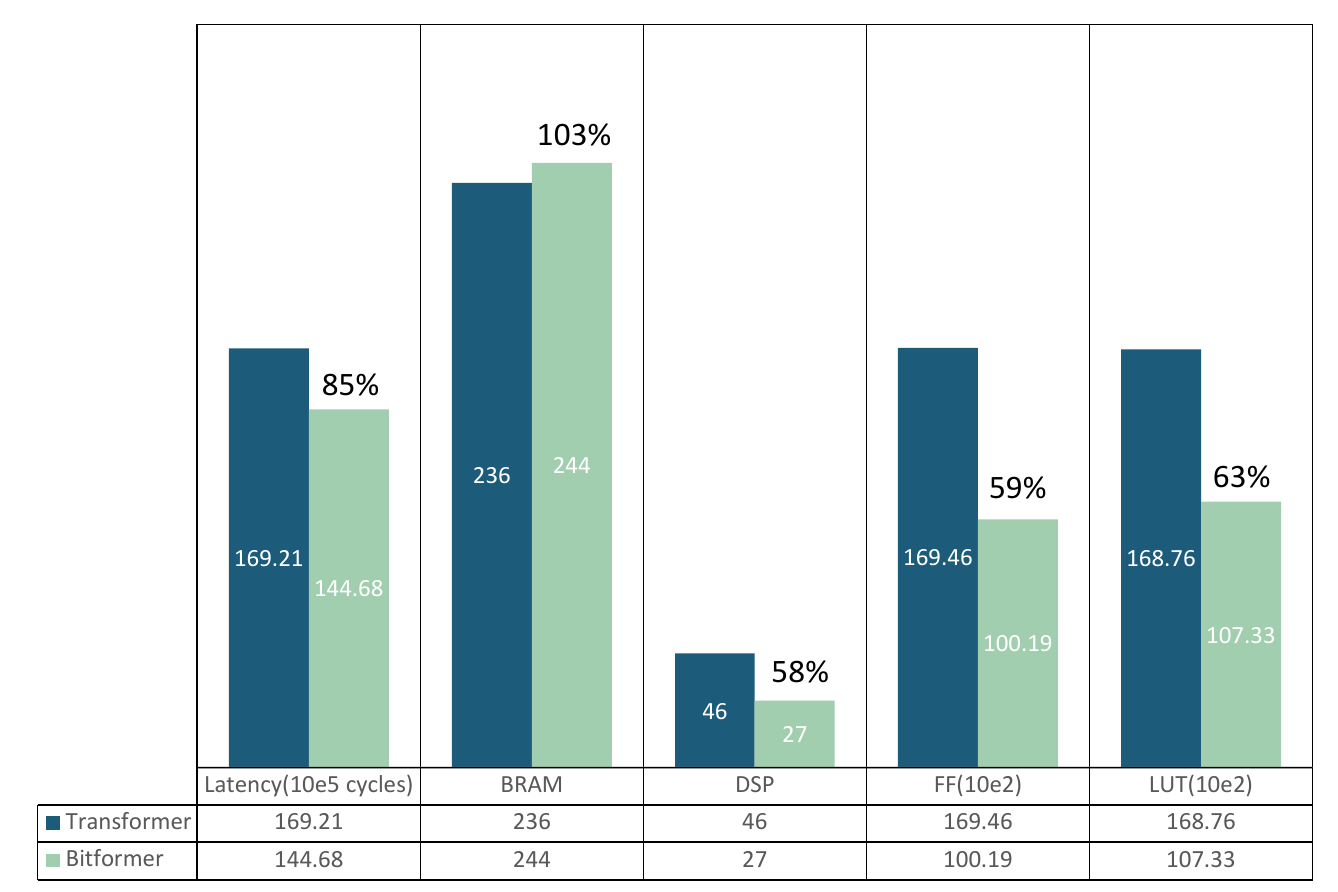}
    \caption{Compare the performance between attention and our bitwise attention on FPGA.}
    \label{fig:FPGA}
\end{figure}
\section{Conclusion}
In this work, we have introduced Bitformer, a groundbreaking Transformer model tailored to tackle the challenges of processing and analyzing extensive data on edge devices. Our primary innovations replace the traditional floating-point multiplication with XOR operations on binary data. These modifications not only maintain the attention mechanism's ability for long-range data analytics but also significantly reduce its computational complexity, making Bitformer exceptionally well-suited for deployment on resource-constrained edge devices, where computational power and energy efficiency are critical considerations.

Despite the radical reduction in computational complexity, Bitformer competes favorably with standard Transformer models in various tasks across NLP and CV, which are frequently encountered in big data analytics scenarios. Our work attests to the effectiveness of Bitformer, thereby establishing that computational efficiency does not need to come at the cost of performance in data analytics tasks.

Bitformer signifies a remarkable advancement in the paradigm of big data analysis within edge device environments. Its proficiency in efficiently managing large-scale data analytics while upholding high performance is a transformative development. Amid the escalating deluge of big data, Bitformer's computational efficiency, combined with its adaptability to edge devices, introduces novel avenues for big data analysis. Consequently, our research presents a promising solution to the challenges encountered in the realm of big data analytics.

Moving forward, while Bitformer's potential has been demonstrated through experiments, our focus will shift towards a more targeted exploration of its capabilities on edge devices. Our future work aims to delve deeper into the realm of software-hardware co-design, aiming to unlock performance optimizations achievable through the binary encoding of core algorithms and personalized hardware-centric enhancements.

In conclusion, Bitformer serves not only as an efficient tool for data analysis but also as an exemplar of the potential synergy between software innovation and hardware efficiency. With its ability to tackle complex data analytics tasks on edge devices, Bitformer contributes significantly to the evolution of big data analytics and paves the way for a new era of efficient computing.

\section{Acknowledgement}
This research was financially supported by the National SKA Program of China under the grant number 2020SKA0120202, as well as the National Natural Science Foundation of China under the grant number U2032125. Prof Chang's research is partly supported by VC Research (VCR 0000212).

%% file: Appendix.tex
\begin{figure}
    \centering
    \includegraphics[width = \linewidth]{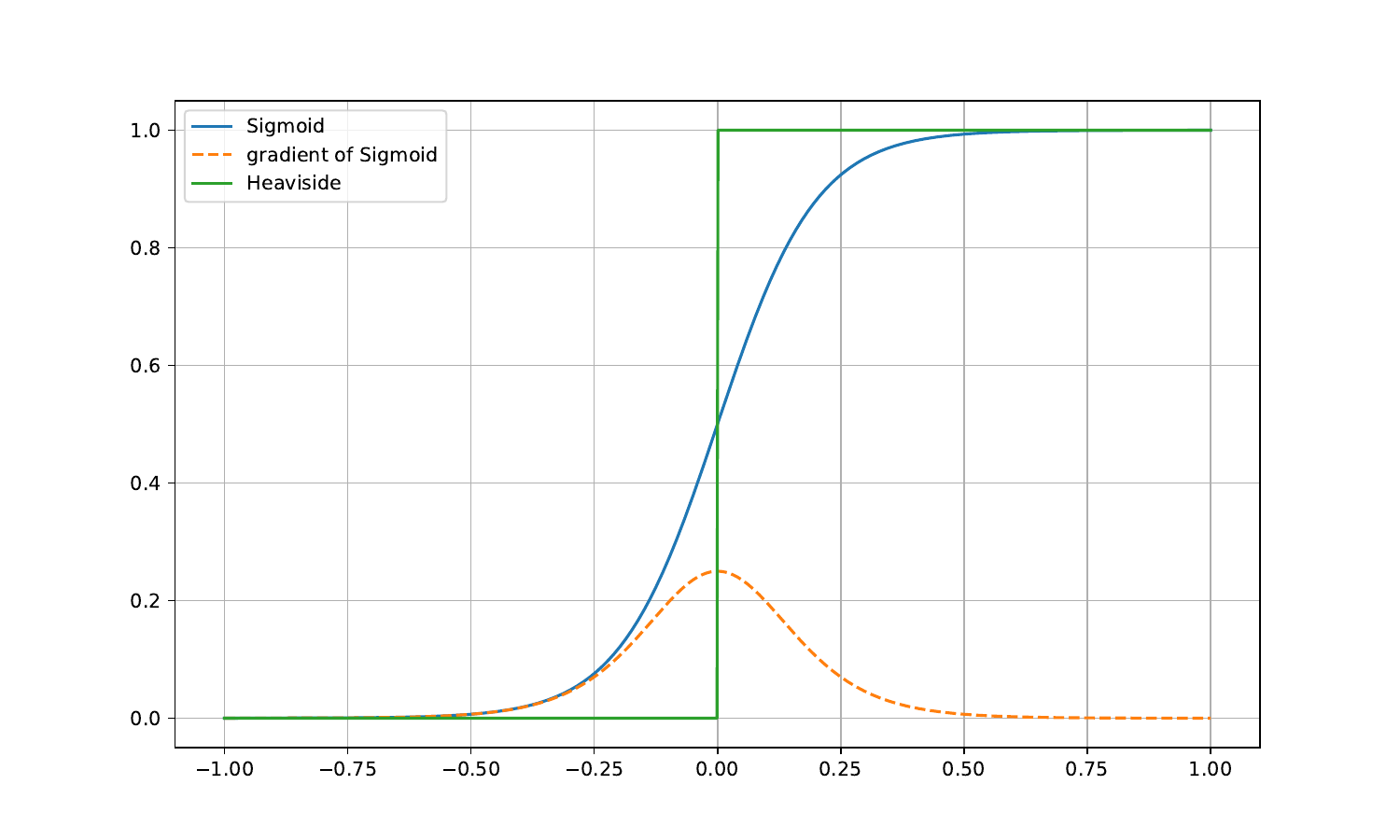}
     \caption{The similiarity between Heaviside and Sigmoid.}
    \label{fig:sigmoid}
\end{figure}
\begin{appendices}
\section{The approximation of the TIF operation}
Consume there is a variant $x$ and time step $T$, we get $x_b$:

\begin{equation}
     x_b = \mathbf{TIF}(x) =\mathbf{IF}(\left [X^1,X^2,..,X^T \right ], T)
\end{equation}
 then we get the sum of $x_b$:
 \begin{equation}
     \mathfrak{S} (x)  = \sum_{i=1}^{T} x_b = \lfloor Tx \rfloor = \lfloor \lambda_0 +\alpha_0 \rfloor  =  \lambda_0
 \end{equation}
 $\lambda_0$ stands for the integer part of $Tx$ and $\alpha_0$ stands for the decimal part which means that $\alpha_0\in[0,1]$. Similarly we have:
  \begin{equation}
     \mathfrak{S} (y)  = \sum_{i=1}^{T} y_b = \lfloor Ty \rfloor =   \lfloor \lambda_1 +\alpha_1 \rfloor  =  \lambda_1
 \end{equation}
 then we have:
\begin{align}
\Gamma_T &= \frac{x}{y} : \frac{\mathfrak{S}(x)}{\mathfrak{S}(y)} = \frac{Tx}{Ty} : \frac{\mathfrak{S}(x)}{\mathfrak{S}(y)} \\
&= \frac{\lambda_0 + \alpha_0}{\lambda_1 + \alpha_1} : \frac{\lambda_0}{\lambda_1} = \frac{1 + \frac{\alpha_0}{\lambda_0}}{1 + \frac{\alpha_1}{\lambda_1}} \\
&= 1 + \frac{\frac{\alpha_0}{\lambda_0} - \frac{\alpha_1}{\lambda_1}}{1 + \frac{\alpha_1}{\lambda_1}}
\end{align}

Since the $\alpha_0,\alpha_1\in[0,1]$ and $\lambda_0 $,$\lambda_0$ increase as the $T$ increase, we have $\lim_{T \to \infty }\Gamma_T = 1$, 
which means that $\forall \  \varepsilon >0$, $\exists \  T_0 >0, \forall T > T_0$, we have 
\begin{equation}
   \left | \frac{x}{y}-\frac{\sum_{i=1}^{T} x_b}{\sum_{i=1}^{T} y_b}\right |  < \varepsilon
\end{equation}
This proved that the TIF operation is a quantifiable conversion from float to binaries data.

\section{Backpropagation}
The deep learning's training process employs backpropagation, a technique that calculates the gradients of the loss function with respect to the model's parameters. These computed gradients are subsequently utilized to update the model parameters, thereby minimizing the loss function. It is crucial to note that all operations involved in the computation of the loss function must be differentiable.

In our unique approach, we incorporate a novel attention mechanism that includes a non-differentiable spike operation and a non-differentiable Hamming distance operation. Despite the non-differentiability of these operations, we can still incorporate them into our models by introducing surrogate gradients at these non-differentiable points.

A surrogate gradient is fundamentally the gradient of a differentiable function that serves as an approximation of the non-differentiable function. 
During the backpropagation process, we substitute the sigmoid function for the Heaviside function, enabling us to complete the backpropagation:
\begin{equation}
    \sigma(x) = \frac{1}{1 + e^{-x}}
\end{equation}
The sigmoid function is differentiable within its domain, which makes it useful for computing gradients during backpropagation. In addition, within the range of [-1,1], the sigmoid function has a similar effect to the Heaviside function. Fig.\ref{fig:sigmoid} demonstrates the similarity between the Heaviside and sigmoid functions.

When it comes to calculating the Hamming distance, we use Mean Square Error(MSE) as an approximation function. It's important to highlight that when dealing with binaries vectors in the computation of the Hamming distance, the value of the Hamming distance is equated to MSE. This equivalence arises from the binaries nature of the vectors involved in the computation:
\begin{equation}
     \phi(a,b)  = \sum_{i=1}^{d} \text{XOR}(a_i,b_i) \\
     =\sum_{i=1}^{d}(a_i-b_i)^2 = d\times \text{MSE}(a,b) 
\end{equation}

\begin{figure}
    \centering
    \includegraphics[width = \linewidth]{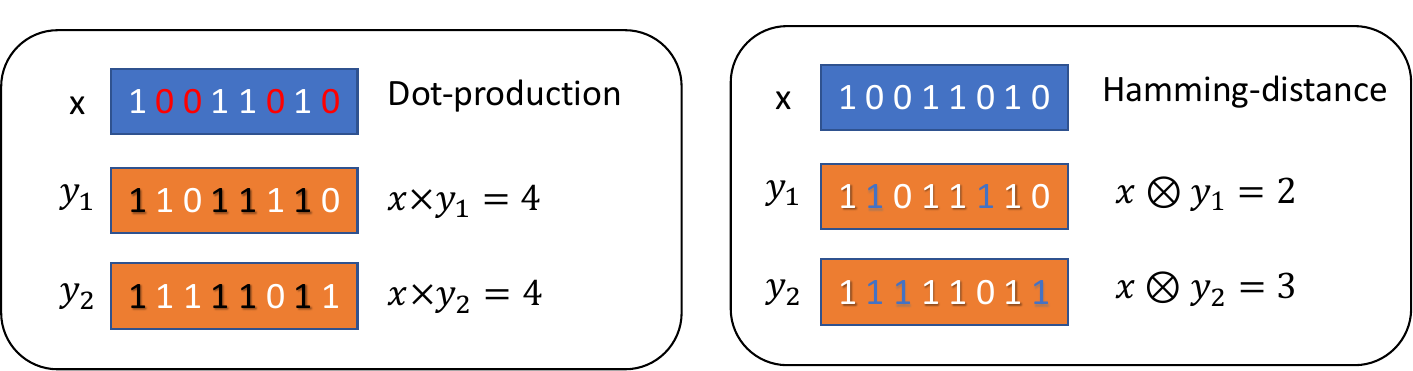}
    \caption{Compare Hamming distance with dot product for binaries data. The nature of binaries data makes it difficult to compare values of 0 effectively. Under the dot product operation, the vectors $y_1$ and $y_2$ lack distinctiveness with respect to $x$. However, the utilization of XOR enables differentiation between the two.}
    \label{fig:compare-distance}
\end{figure}
\section{Hamming Distance}
When comparing similarities in binaries data, the Hamming Distance offers more benefits compared to the dot product. 
In the case of float data, calculating the dot product of two float vectors is an effective and intuitive method, which was also utilized in the original Transformer. However, this approach fails to accurately capture the similarity between two vectors in binaries data since the positions where the values are 0 do not effectively discern the differences in other vectors. Fig.\ref{fig:compare-distance} shows an example to reveal the difference between the two methods of describing distance in binaries data.

\end{appendices}